\documentclass[letterpaper, 10 pt, conference]{ieeeconf}
\bstctlcite{bstctl:etal, bstctl:nodash, bstctl:simpurl}
\pdfoutput=1
\IEEEoverridecommandlockouts
\usepackage{graphicx}
\usepackage{subcaption}
\usepackage{amsmath}
\usepackage{multirow}
\usepackage{booktabs}
\usepackage[percent]{overpic}
\usepackage{adjustbox}
\usepackage{amssymb}
\makeatletter
\let\NAT@parse\undefined
\DeclareRobustCommand\onedot{\futurelet\@let@token\@onedot}

\usepackage[colorlinks,pagebackref=false,citecolor=blue,bookmarks=false,hypertexnames=true]{hyperref}

\ifx01 
\usepackage{ulem}
\newcommand{\del}[1]{\textcolor{red}{$<$\sout{#1}$>$}}
\newcommand{\add}[1]{\textcolor{blue}{$<$#1$>$}}
\else 
\newcommand{\del}[1]{}
\newcommand{\add}[1]{#1}
\fi

\title{\LARGE \bf
A Hybrid Sparse-Dense Monocular SLAM \del{framework}\add{System}\\ for Autonomous Driving
}

\author{Louis Gallagher$^{1}$, Varun Ravi Kumar$^{2}$, Senthil Yogamani$^{3}$ and John B. McDonald$^{1}$
\thanks{*Research presented in this paper was supported by IRC grant GOIPG/2016/1320 HEA COVID extension fund, Science Foundation Ireland grant 13/RC/2094 to Lero - the Irish Software Research Centre and Science Foundation Ireland grant 16/RI/3399.}
\thanks{$^{1}$Louis Gallagher and John McDonald are with Lero - the Irish Software Research Centre and the Department of Computer Science, Maynooth University, Maynooth, Ireland
{\tt\small \{Louis.Gallagher, John.McDonald\}@mu.ie}}%
\thanks{$^{2}$Varun Ravi Kumar is with Valeo DAR Kronach, Germany. 
{\tt\small varun.ravi-kumar@valeo.com}}%
\thanks{$^{3}$Senthil Yogamani is with Valeo Vision Systems, Ireland
        {\tt\small senthil.yogamani@valeo.com}
\newline 978-1-6654-1213-1/21/\$31.00 \textcopyright 2021 IEEE}
}
\begin{document}

\bstctlcite{IEEEexample:BSTcontrol}

\maketitle
\thispagestyle{empty}
\pagestyle{empty}
\begin{abstract}
In this paper, we present a system for incrementally reconstructing a dense $3D$ model of the geometry of an outdoor environment using a single monocular camera attached to a moving vehicle. Dense models provide a rich representation of the environment facilitating higher-level scene understanding, perception, and planning. Our system employs dense depth prediction with a hybrid mapping architecture combining state-of-the-art sparse features and dense fusion-based visual SLAM algorithms within an integrated framework. Our novel contributions include design of hybrid sparse-dense camera tracking and loop closure, and scale estimation improvements in dense depth prediction. We use the motion estimates from the sparse method to overcome the large and variable inter-frame displacement typical of outdoor vehicle scenarios. Our system then registers the live image with the dense model using whole-image alignment. This enables the fusion of the live frame and dense depth prediction into the model. Global consistency and alignment between the sparse and dense models are achieved by applying pose constraints from the sparse method directly within the deformation of the dense model. We provide qualitative and quantitative results for both trajectory estimation and surface reconstruction accuracy, demonstrating competitive performance on the KITTI dataset. Qualitative results of the proposed approach \del{is}\add{are} illustrated in \url{https://youtu.be/Pn2uaVqjskY}.\add{ Source code for the project is publicly available at the following repository \url{https://github.com/robotvisionmu/DenseMonoSLAM}}
\end{abstract}
\section{Introduction}

    Over the past decade, approaches to fusion-based dense visual SLAM have demonstrated the ability to build high fidelity dense $3D$ models of an environment facilitating higher-level scene understanding, perception and planning \cite{kinectfusion}, \cite{kintinuous}, \cite{elasticfusion}, \cite{bundlefusion}. However, given that these techniques typically require active RGB-D sensors, their applicability in outdoor scenarios has been limited. On the other hand, sparse and semi-dense monocular systems have found a large degree of success in these environments but are limited in their resulting sparse or partially dense representations \cite{orbslam}, \cite{orbslam}, \cite{orbslam3}, \cite{orb_vi}, \cite{lsd_slam}, \cite{svo}, \cite{svo2}. Here, we investigate the potential of recent results in monocular dense depth prediction in combining these approaches' strengths. In particular, we employ dense depth prediction in developing a monocular SLAM system that achieves comparable accuracy to sparse camera tracking algorithms while allowing dense fusion-based modeling of the environment. We demonstrate the effectiveness of this approach by tracking and mapping an environment from a single monocular camera attached to a moving vehicle. \par
\begin{figure}[!t]
  \centering
\begin{tabular}{cc}
 	\begin{overpic}[trim={0.4cm 0 0 0},clip,width=0.498\columnwidth]
 	{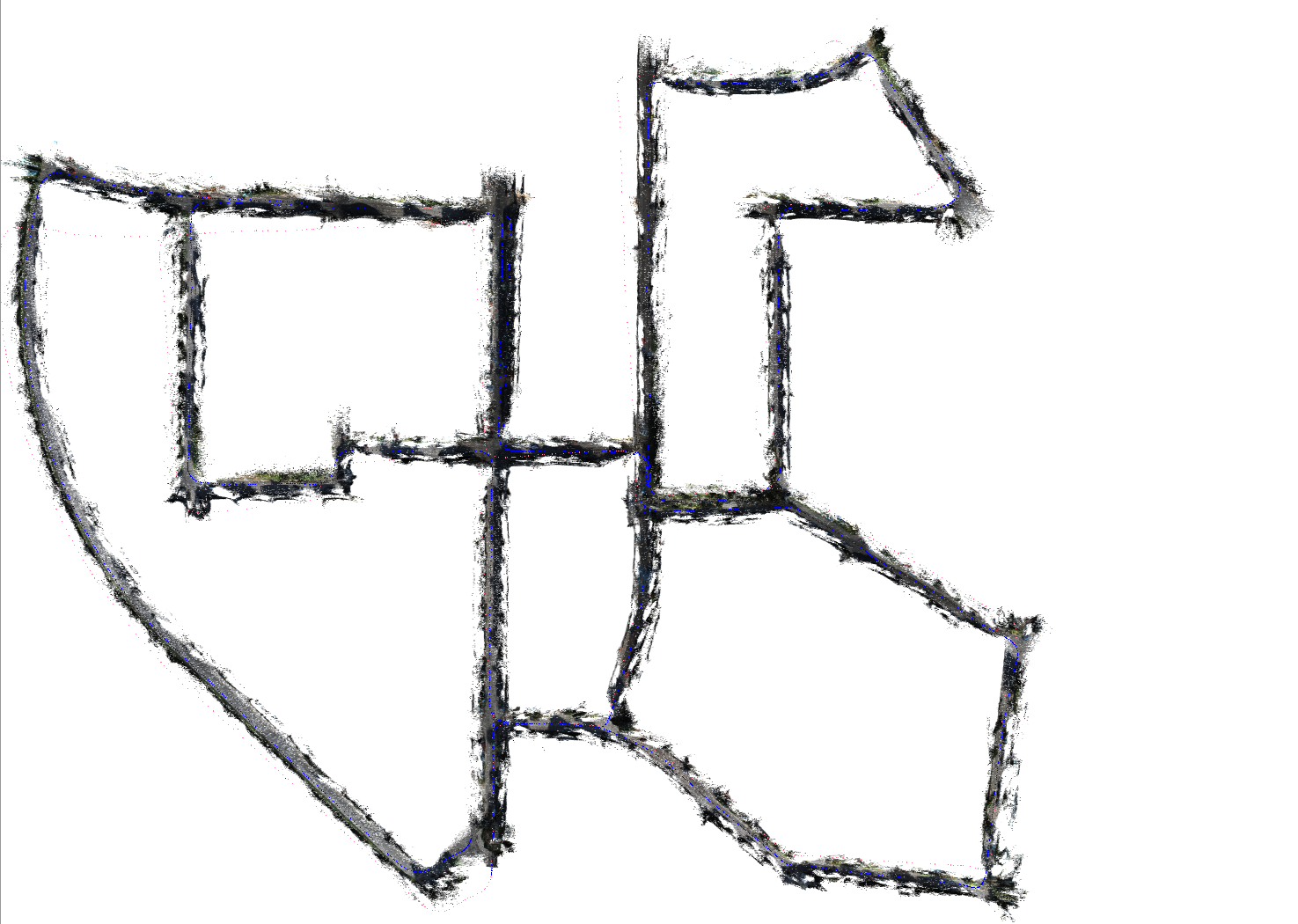}
    \put (0,3) {{$\displaystyle\textcolor{red}{\text{(a)}}$}}
    \end{overpic}
    \begin{overpic}[width=0.498\columnwidth]{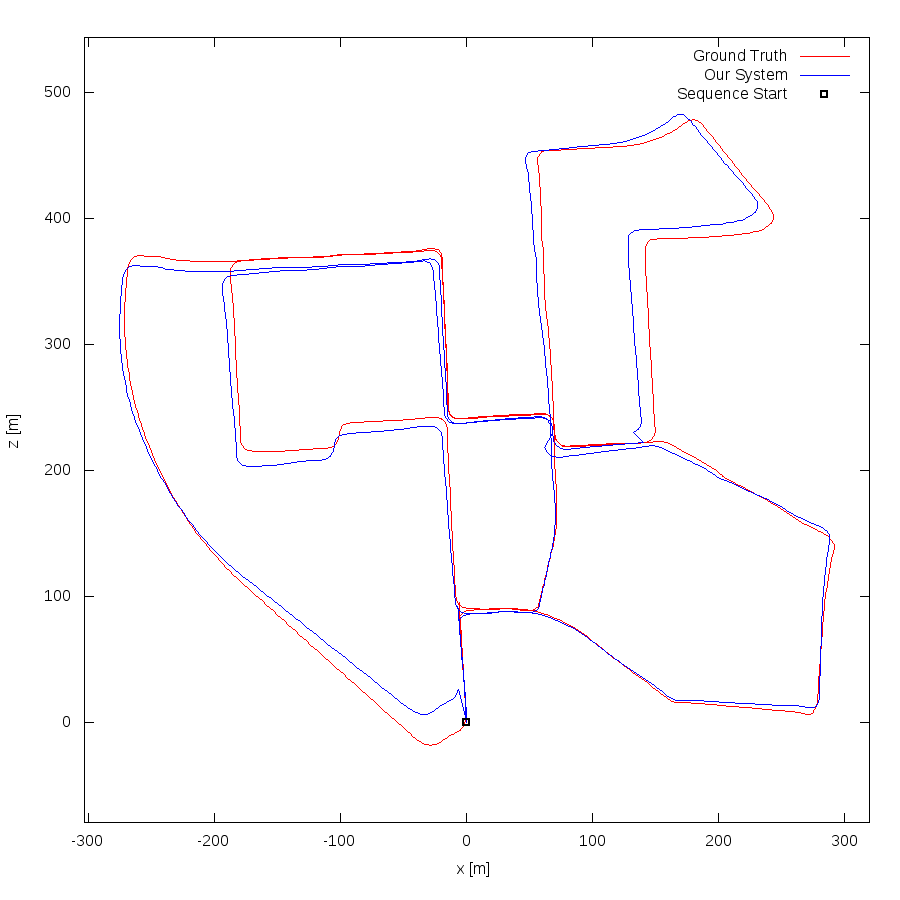}
    \put (0,3) {{$\displaystyle\textcolor{red}{\text{(b)}}$}}
    \end{overpic} \\
    \begin{overpic}[trim={0.4cm 0 0 0},clip,width=\columnwidth]
    {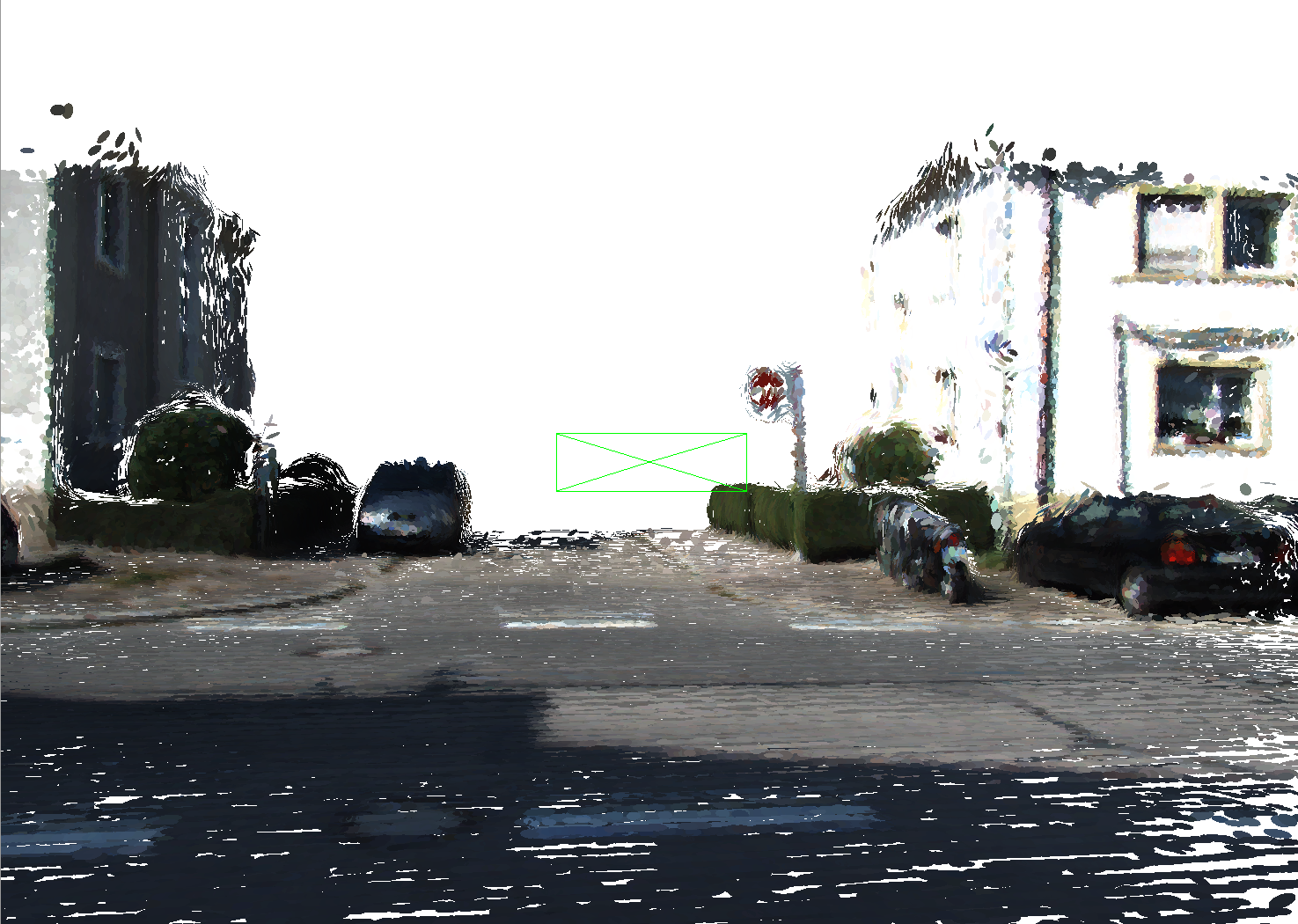}
    \put (0.5,2) {{$\displaystyle\textcolor{green}{\text{(c)}}$}}
    \end{overpic}
\end{tabular}
   \caption{\textbf{Trajectory and dense reconstruction of an odometry sequence from \cite{kitti_benchmark, kitti_raw}.} (a) shows the final surfel model produced by our system. The model contains approximately $16m$ surfels. (b) shows the estimated trajectory \del{(projected onto the $xy$ plane)} alongside the ground truth trajectory. (c) shows an up-close view of the reconstruction as the vehicle passes through an intersection contained in (a).}
   \label{fig1}
\end{figure}

Using passive vision sensors over active sensors, such as LiDAR, in autonomous driving SLAM systems has many advantages. A calibrated monocular camera is inexpensive, lightweight, and can be used as a direction sensor providing rich photometric and geometric measurements of the scene at every frame. In contrast, LiDAR is expensive, heavy, and produces a relatively sparse signal compared to the dense measurements provided by a vision sensor. Indeed, recent deep learning-based depth estimation and object detection indicates that the accuracy advantage of LiDAR over monocular vision-based systems is closing \cite{lidar}.\par

On the other hand, monocular vision systems suffer from ambiguity and drift in the estimated scale of the reconstruction \cite{scaledrift-aware} and are not as robust as stereo and multi-sensor systems in texture-less regions and during fast camera motion \cite{orbslam3, orb_vi}. Furthermore, with the exception of a few techniques for dense and semi-dense map representations \cite{dtam, lsd_slam, semi_dense_odom}, these systems produce sparse reconstructions which only offer a coarse-grained description of the scene's geometry \cite{monoslam, ptam, graphslam, orbslam, orbslam2}.\par 

Multi-sensor platforms attempt to leverage the combined advantages of monocular, inertial, stereo, LiDAR, and GPS sensors. However, such platforms need careful internal and external calibration, which can be done online to maintain synchronization and geometric alignment between the sensors and, ultimately, achieve accurate performance. In the autonomous driving domain, even the most carefully calibrated rig will move and warp in real-time due to the effects of heat, wear and tear of mechanical parts like tires, variations in loading of the vehicle, and the knocks and bumps inherent in driving on roads \cite{stereo_continuous_calibration}. GPS comes with its complexities, requiring line-of-sight with at least four satellites to infer position, which can prove challenging in urban environments \cite{gps_solutions}.\par

More recently, several systems have emerged that represent the state-of-the-art in traditional monocular SLAM, and visual-odometry performance \cite{orbslam3, orbslam2, dso, ldso, svo, svo2, lsd_slam}. The ORB-SLAM open-source SLAM frameworks \cite{orbslam,orbslam2,orbslam3} in particular represent the state-of-the-art in monocular SLAM. Similar to PTAM, ORB-SLAM \cite{orbslam2} splits the SLAM problem into sub-problems that can be solved in different threads; (i) a feature-based camera tracking thread that runs in real-time; (ii) a local mapping thread that takes keyframes extracted by the first thread and optimizes a local area of the map around this keyframe using bundle-adjustment (BA); and (iii) a global mapping thread that integrates loop closure constraints and maintains global consistency.\par

\del{A class of v}\add{V}isual SLAM systems based around the concepts of dense alternation, every pixel fusion, and direct whole-image alignment-based camera tracking, have shown that it is possible to build dense, high-fidelity models of indoor scenes in real-time with active depth sensors such as the Microsoft Kinect\del{\cite{kinectfusion, elasticfusion, kintinuous, bundlefusion, badslam}}. \add{These systems and the models they produce have many benefits in downstream machine perceptions and spatial reasoning tasks.}\del{When it comes to downstream machine perception, and spatial reasoning tasks, these systems and the models they produce have many benefits.} They can produce `watertight' maps that model the continuous surface that underlies the scene. For example, this allows them to synthesize accurate novel views and realistically fuse augmentations into the scene for AR effects \cite{kinectfusion, elasticfusion, kintinuous, bundlefusion, badslam}. High-resolution dense models may become more necessary for robot planning in the future to accommodate a broader range of scenarios (e.g., analyzing surface geometry in unevenly paved roads \cite{dense_path_planning}, obstacle detection and avoidance \cite{lidar}, etc.).\par

More and more deep learning is being used to improve the performance of SLAM systems. Broadly speaking, neural networks have been applied in two critical areas within \del{SLAM}\add{the SLAM problem}; in the representation of geometric and visual information \cite{codeslam, imap, representation_learning}, and in overcoming the ill-posedness of many of SLAM's subproblems, for example, depth estimation from monocular images \cite{unrectdepthnet, depth_ego_unsup, depth_eigen, left_right_consistency}. \del{Several systems have been presented that demonstrate hybrid approaches to monocular visual SLAM and VO \cite{codeslam, cnn_slam, d3vo}, while other approaches implement a fully learned SLAM pipeline \cite{deep_tam, deep_vo}. In comparison to this work, these systems are either not suited to fast, outdoor camera motion \cite{codeslam, deep_tam, cnn_slam} or do not produce dense models \cite{d3vo, deep_vo}}\add{Several systems have been presented that demonstrate hybrid approaches to monocular visual SLAM and VO \cite{deepfactors, codeslam, cnn_slam, d3vo}, while other approaches implement a fully learned SLAM pipeline \cite{deep_tam, deep_vo}. In comparison to this work, these systems have either not been designed with large-scale fast, outdoor camera motion in mind \cite{deepfactors, codeslam, deep_tam, cnn_slam}, or do not produce dense models \cite{d3vo, deep_vo}.}\par

\add{The work most similar to \add{our aproach}\del{the current work} is the dense surfel-based fusion system of \cite{scalable_dense_surfel}, \del{in which a hybrid approach is proposed that combines the}\add{combining} camera tracking from a sparse SLAM system (ORB-SLAM2 \cite{orbslam2}) with a dense fusion-based back-end. The substantial difference with the current work is \del{in} the degree to which they couple the sparse and dense maps. In \cite{scalable_dense_surfel}, \add{a sub-mapping approach is employed where} each surfel is anchored to a reference keyframe from the sparse system. Local and global optimisations of the keyframe poses by the sparse system are reflected in the dense map by ensuring the pose of a surfel relative to its anchor keyframe remains constant. In contrast, our system maintains a much looser coupling between the sparse and dense representations. The sparse pose is used to initialise a dense alignment that registers the camera to the dense map allowing for fusion of the current frame. Loop closures from the sparse system are used to constrain a deformation graph that non-rigidly deforms the surfels themselves. In this way, the benefits of each system are leveraged, while the systems themselves are only indirectly connected through $6$-DOF transformations and $3D$ constraints. \add{Furthermore, w}e quantitatively demonstrate the effectiveness of our approach in outdoor automotive environments using deep-learned depth estimation.}\par

\del{The contribution of this paper is the development of a system for live dense metric $3D$ reconstruction in outdoor environments using a single RGB camera attached to a moving vehicle. Although other researchers have reported semi-dense SLAM or dense VO approaches, this is the first fully dense SLAM system evaluated on automotive scenarios to the best of our knowledge. The novelty of our approach lies in the use of a dense depth prediction network \cite{unrectdepthnet} within a hybrid architecture combining state-of-the-art sparse feature tracking and dense fusion-based visual mapping algorithms. We improve our previous approach to dense depth prediction \cite{unrectdepthnet} for the SLAM application by the addition of new regularization losses and better scale estimation.}
\add{The contribution of this paper is the development of a system for live dense metric $3D$ reconstruction in outdoor environments using a single RGB camera attached to a moving vehicle. Although other researchers have reported dense and semi-dense SLAM or dense VO approaches, to the best of our knowledge, this is the first fully dense SLAM system \emph{quantitatively} evaluated on automotive scenarios. The novelty of our approach lies in the use of a dense depth prediction network \cite{unrectdepthnet} within a hybrid architecture, combining state-of-the-art sparse feature tracking and dense fusion-based visual mapping algorithms in a loosely coupled fashion. We improve our previous approach to dense depth prediction \cite{unrectdepthnet} for the SLAM application by the addition of new regularization losses and better scale estimation.}

Using the motion estimates from the sparse method to overcome the large and variable inter-frame displacement typical of outdoor vehicle scenarios, our system then registers the live image with the dense model using whole-image alignment. This allows dense depth prediction and fusion of the live frame into the model. Global consistency and alignment between the sparse and dense models are achieved by applying pose constraints from the sparse method directly within the deformation of the dense model. We evaluate the method over the KITTI benchmark dataset, providing qualitative and quantitative results for both the trajectory and surface reconstruction accuracy.\par

\section{A Hybrid Approach}
\begin{figure*}[!t]
    \centering
    \includegraphics[width=\textwidth]{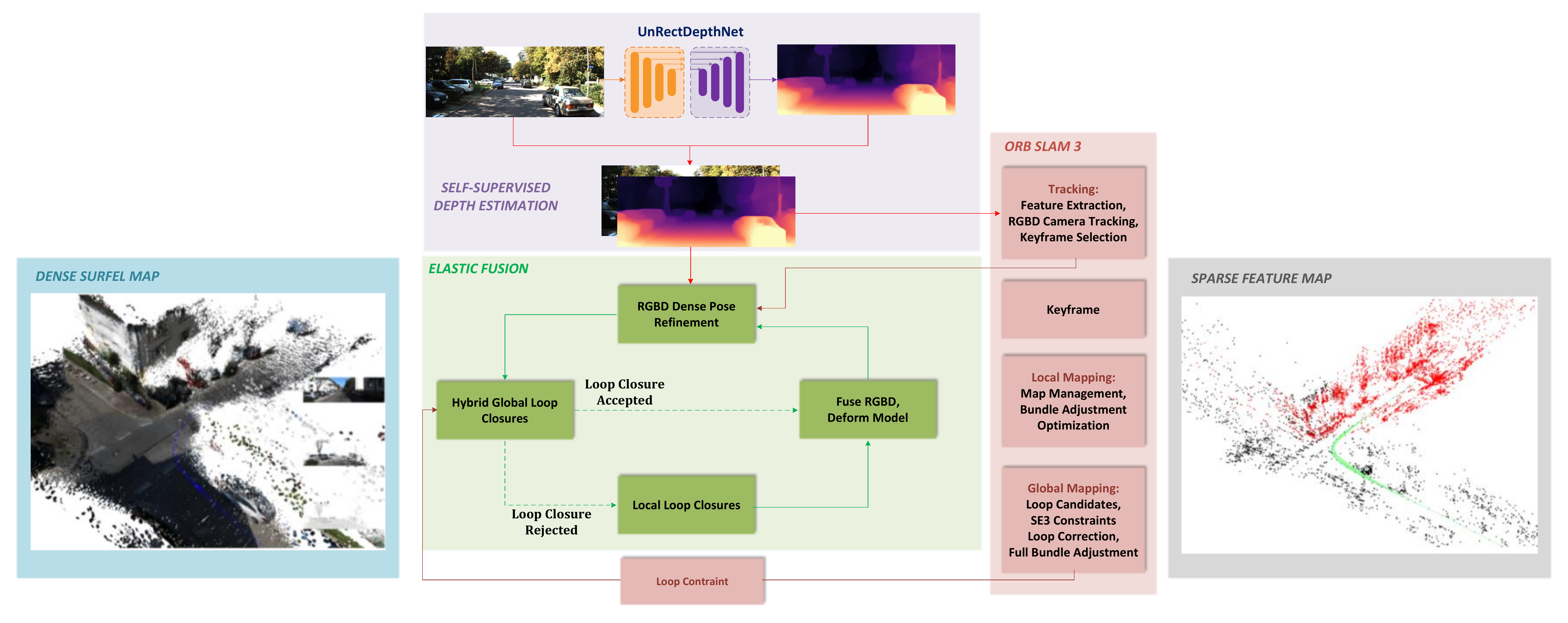}
    \caption{\add{\textbf{An overview of our system architecture.}} UnRectDepthNet is used to estimate a depth map for the current live camera image. ORB-SLAM then tracks the motion of the camera since the last frame. Internally ORB-SLAM continues to extract keyframes and pass them to its local and global mapping threads. The sparse feature map on the right is used internally by ORB-SLAM. Once the camera motion has been estimated, the pose is used to synthesize a dense view of the current active surfel model. The pose estimate is further refined by performing a direct joint photometric and geometric alignment between the synthetic and live RGB-D frames, taking the pose estimate from ORB-SLAM as an initialization point. Assuming no local or global loop closures have occurred during this timestep, the live RGB image and predicted depth are fused into the surfel model. If ORB-SLAM has detected a global loop closure, then a global deformation is attempted. Surface-to-surface constraints for optimization of the deformation graph are generated from the pre-corrected and corrected (post loop closure) poses of the camera $P_{kf}^t$ and $\hat{P}_{kf}^t$ respectively.
    If deformation graph optimization fails or no global loop is detected by ORB-SLAM, a local loop closure is attempted as per the original EF algorithm.}
   \label{fig:arch}
\end{figure*}

We take a hybrid approach to dense monocular tracking and mapping. A feature-based SLAM system, ORB-SLAM3\cite{orbslam3}, is used to provide an initial estimate of the camera's pose for each frame. The pipeline then follows a dense alternation architecture, extending the ElasticFusion (EF) SLAM system \cite{elasticfusion}, in which the map is first held fixed. At the same time, the camera is tracked against it, using the initial pose estimate from the previous step. Once the camera pose has been estimated, it can fuse the current frame into the map. We densely predict per pixel metric depth estimates using a SOTA self-supervised convolutional neural network, UnRectDepthNet \cite{unrectdepthnet}. The various sub-systems predominantly operate on different processors; the GPU is used extensively by the dense alternation and depth prediction network, while ORB-SLAM operates on the CPU. Our hybrid architecture is summarised as follows:
\begin{enumerate}
  \item A scale-aware depth prediction network is used to estimate a metric depth map for each frame. An initial estimate of the camera motion is made using ORB-SLAM's feature-based camera tracker, which is suited to the fast motion of the \del{car}\add{vehicle}.
  \item The initial pose estimate is further refined by aligning it to the current active model in view of the camera. 
  \item Live RGB images and corresponding predicted metric depth maps are fused into a global dense surfel model of the scene. The surfel model is divided into an active and an inactive portion as per the original EF algorithm.
  \item When ORB-SLAM identifies a loop closure, we use the resulting loop closure constraint within the EF deformation graph to correct the geometry of the dense surfaces. This brings the previously visited inactive portion of the map back into alignment with the current active portion. Importantly, this also keeps the different map and camera trajectories of ORB-SLAM and EF consistent.
\end{enumerate}
Figure \ref{fig:arch} shows the architecture of our system. In the remainder of this section, we will describe how these key elements are combined into one system that builds consistent dense surfel models from a monocular camera attached to a moving vehicle. The section provides complete coverage of the system where the principal novelty is covered in \ref{sec:depth}, \ref{sec:hybridcamera}, and \ref{sec:hybridloop}. We then present and discuss experimental results that demonstrate the efficacy of our system in \ref{sec:evaluation}.  

\subsection{Scale-aware depth estimation}
\label{sec:depth}

Following UnRectDepthNet \cite{unrectdepthnet}, we establish the same \emph{structure-from-motion (SfM)} framework for self-supervised depth estimation. We carry out the view synthesis by employing the pinhole camera projection model. The final objective consists of a photometric term $\mathcal{L}_p$ and an edge smoothness regularization term $\mathcal{L}_s$. In addition, the cross-sequence depth consistency loss $\mathcal{L}_{dc}$ and the scale recovery approach are employed. In the following paragraphs, we discuss the new improvements that lead to a considerable gain in accuracy.

We employ\del{ed} feature-metric losses from \cite{shu2020featdepth}, where discriminative, $\mathcal{L}_{dis}$, and convergent, $\mathcal{L}_{cvt}$, losses are calculated on the current frame's, $I_t$'s, feature representation by incorporating a self-attention autoencoder to obtain robust global features of the scene. The primary goal of $\mathcal{L}_{dis}$ and $\mathcal{L}_{cvt}$ is to keep the optimization objective from getting trapped at several local minima for low texture areas such as sky and road. It is a crucial loss feature that uses image gradients to penalize small slopes while emphasizing low-texture areas. 
The self-supervised loss landscapes are restricted from forming proper convergence basins using first-order derivatives to regularize the target features.
However, enforcing discriminative loss alone will not ensure that we reach the best solution during gradient descent. As there exists inconsistency among first-order gradients, (\textit{i.e.}, gradients that are spatially adjacent point in opposite directions), convergent $\mathcal{L}_{cvt}$ loss is employed
to enable gradient descent from a far-off distance. It has a relatively large convergence radius and expresses the loss to have uniform gradients throughout the optimization step by supporting feature gradient smoothness and large convergence radii accordingly. The total objective $\mathcal{L}_{depth}$ is
\begin{align}
    \mathcal{L}_{depth} &= \mathcal{L}_r(I_t,\hat{I}_{t'\to t}) + \alpha~\mathcal{L}_s(\hat{D}_t) + \rho~\mathcal{L}_{dc}(\hat{D}_t,\hat{D}_{t'}) \\
    &+ \mathcal{L}_r(\hat{F}_t,\hat{F}_{t'\to t}) + \zeta~\mathcal{L}_{dis}(I_t, \hat{F}_t) 
    + \eta~\mathcal{L}_{cvt}(I_t, \hat{F}_t) \nonumber
\end{align}
\add{Where $\hat{F}_t$ is the estimated feature at time $t$, $I_t$ is the current colour image, $\hat{D}_t$ the estimated depth map,} $\mathcal{L}_r$ is the standard reconstruction matching term, and $\alpha$, $\rho$, $\zeta$ and $\eta$ weigh the smoothness term $\mathcal{L}_s$, cross-sequence depth consistency $\mathcal{L}_{dc}$, discriminative $\mathcal{L}_{dis}$ and convergent $\mathcal{L}_{cvt}$ losses respectively.\par

\del{Scale ambiguity is a challenging problem in monocular depth estimation. Many authors~\cite{garg2016unsupervised, zhou2017unsupervised} would note that the problem of estimating depth from a single RGB image is an ill-posed inverse problem when researching monocular depth estimation. \textit{i.e}, several 3D scenes observed in the world may also correspond to the same 2D plane. This does not preclude us from making specific predictions, but it does suggest that all depth values would be relative to one another. Such scale variation can severely affect the SLAM system.}\par
 \del{As a result, an absolute value is needed to serve as an anchor point, provided through a measurement from another dedicated sensor to obtain actual depth estimation. We use a combination of Velodyne point cloud and calibration information to improve the estimation of scale. We use Velodyne Lidar as ground truth and estimate the scale factor by associating the calculation with the correct pixel in the image plane, minimizing $\ell_{2}$ loss. In addition, we assume depth consistency across training and testing datasets. It can be helpful in datasets with high pose variability, such as KITTI, where the camera is \textit{still} at the same height and looking at the ground from the same perspective. We fit a ground plane and use calibration information and known camera height to find the scale factor. We use a combination to improve our scale estimation.}
 \add{Scale ambiguity is a challenging problem in monocular depth estimation ~\cite{garg2016unsupervised, zhou2017unsupervised}. As a result, an absolute value is needed to serve as an anchor point, provided through a measurement from another dedicated sensor to obtain actual depth estimation. We use a combination of Velodyne point cloud and calibration information to improve the estimation of scale. We use Velodyne Lidar as ground truth and estimate the scale factor by associating the calculation with the correct pixel in the image plane, minimizing $\ell_{2}$ loss. In addition, we assume depth consistency across training and testing datasets. It can be helpful in datasets with high pose variability, such as KITTI, where the camera is \textit{still} at the same height and looking at the ground from the same perspective. We fit a ground plane and use calibration information and known camera height to find the scale factor. We use a combination to improve our scale estimation.}\par
\subsection{ORB-SLAM3 - Feature-based RGBD Tracking}

\del{ORB-SLAM3 \cite{orbslam3} facilitates operation with several sensor configurations,}\del{ including monocular, stereo, RGB-D, and combined visual-inertial inputs.,} \del{\add{however}, here we will focus on its RGB-D operation as it is the mode used within our system.}\add{ ORB-SLAM3 \cite{orbslam3} builds a sparse map of a scene, represented using a \emph{covisibility}} graph where each node in the graph corresponds to a keyframe consisting of a pose and a set of $3D$ points. When two keyframes view common map points, an edge between them is added to the graph. The system has $3$ main threads. The camera tracking thread receives RGB-D frames from the camera, extracts ORB features\del{from it}, and computes an initial pose estimate via motion only BA with the previous frame. The estimate is further refined by aligning the current frame to a local map of covisible keyframes. New keyframes are detected and sent to a background thread. Here the local map in the vicinity of the new keyframe is optimized with a full bundle adjustment. Loops between the latest keyframe and historic keyframes are detected with DBoW\cite{dbow2}. If geometric alignment between the new keyframe and the matched keyframe succeeds, then a loop is closed, and an edge between the two keyframes is added to the graph. The local map in the vicinity of the keyframe is rigidly transformed into place. The rest of the keyframe graph is corrected using pose graph optimization. A final full BA is performed to recover the MAP estimate of all keyframe poses and structure.\par 
\begin{figure*}[t!]
    \centering
    \includegraphics[width=\textwidth]{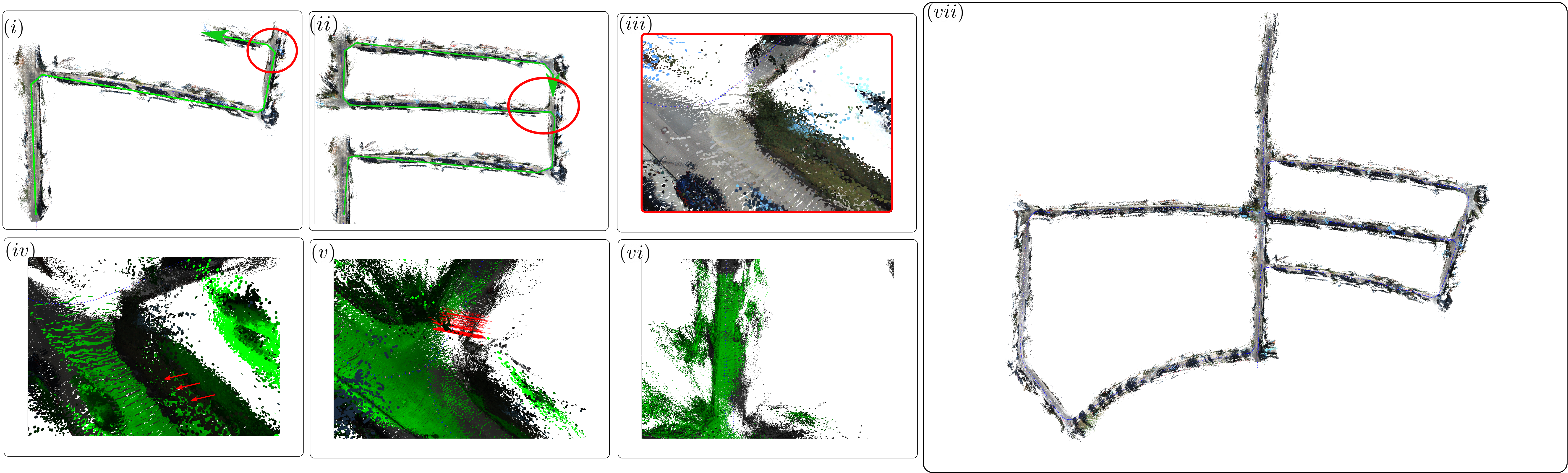}
    \caption{\textbf{A loop closure example}; $(i)$ a car begins exploring (green arrow); $(ii)$ after a period of time a loop is closed between an area previously mapped by the car and its current location; $(iii)$ before a loop closure is applied, drift is evident in the model; $(iv)$ the active portion of the model (green) has drifted from the inactive portion of the model (gray). Red arrows highlight corresponding points between the active and inactive map; $(v)$ a global surface deformation has been applied bringing the active portion of the map back into alignment with the inactive portion. Red lines are surface-to-surface constraints generated from the loop closing pair $P_{kf}^t \rightarrow \hat{P}_{kf}^t$; $(vi)$ as the car moves back into the region of the road already mapped in the first pass, local loop closures are triggered, reactivating portions of the map for use in mapping and tracking\del{(green)}; $(vii)$ the final global model.}
    \label{fig:loop}
\end{figure*}
\subsection{ElasticFusion - Dense Surfel Fusion}

ElasticFusion \cite{elasticfusion} estimates a dense surfel map $\mathcal{M}$ of a scene as viewed from a handheld RGB-D sensor. The map $\mathcal{M}$ is a flat list structure containing a set of surface elements $\mathcal{M}^s$ describing local planar patches of the underlying scene surface. Each $\mathcal{M}^s$ contains a $3D$ position, a normal, a radius $r$ indicating its extent, a confidence value indicating quality of its estimation, an RGB color, a timestamp $t$ when it was last fused, and a timestamp $t_o$ when it was inserted into the map. $\mathcal{M}$ is itself divided into two disjoint subsets; an active portion containing surfels that have recently been measured by the live camera data and an inactive portion containing those surfels that have not been measured in a threshold amount of time. \del{Frame-to-model camera tracking is performed by aligning the current live camera view to a synthetic view of the model rendered using the camera's pose at the previous timestep. Alignment is performed via a joint photometric and geometric non-linear least-squares cost optimization.} \add{Camera tracking is performed by direct alignment of the current camera frame to a synthetic model view using the pose of the camera at the previous timestep.} Frames are fused by back projecting each pixel to a $3D$ point using the depth map and the camera's intrinsic matrix\del{$K$}.\del{If a live point intersects an active surfel in $\mathcal{M}$, they are fused by averaging their properties together.} The system also has a visual place recognition loop closure mechanism based on fern encoding; however, we do not utilize this module. Local loops are identified by aligning views of the active and inactive portions of $\mathcal{M}^s$ in the current camera view, allowing inactive surfels to be reactivated. Both global and local loop closures provide surface-to-surface constraints to optimize a space deforming graph, which is then applied to $\mathcal{M}^s$. Camera tracking and deformation-based loop closures will be discussed in later sections as we describe how we combine them with the other systems.\par

\subsection{Hybrid Camera Tracking}
\label{sec:hybridcamera}

In our system, at each timestep, $t$, the current live camera frame $\mathcal{F}_t$, consisting of color image $\mathbf{I}_t$ and predicted depth map $\mathbf{D}_t^p$, is passed to ORB-SLAM to \del{get}\add{compute} an initial estimate of the camera pose $\mathbf{P}_t$. Note that this is the camera pose before any refinement by the local mapping thread in the case that $\mathcal{F}_t$ is extracted as a keyframe. Though this initial pose is accurate (as can be seen in Table \ref{table:kitti_accuracy}), it cannot be used directly for the fusion of $\mathcal{F}_t$ into $\mathcal{M}$. ORB-SLAM continues to improve its local map, while the dense surfel map does not undergo any significant correction until a local or global loop closure is triggered. This can lead to $\mathbf{P}_t$, and therefore the live frame, becoming misaligned from the dense map. Our goal is to track the camera \emph{and} to build an accurate model, and so we need to balance the accuracy of $\mathbf{P}_t$ against the need to stay aligned with the dense model in order to fuse $\mathcal{F}_t$ accurately. 

To bring the camera back into alignment with the model, we perform a frame-to-model refinement of $\mathbf{P}_t$. The active map around $\mathbf{P}_t$ is rendered into a virtual camera positioned at $\mathbf{P}_t$. A $6$DOF transformation $T \in SE_3$ that aligns the live frame to the virtual one is then estimated in an iterative non-linear least-squares \add{joint photometric and geometric} optimization embedded in a $3$ level image pyramid. Composing $T$ with $\mathbf{P}_t$ yields a refined pose estimate for the current frame that can now be used to fusion as per the original EF algorithm \cite{elasticfusion}.\par
\subsection{Hybrid Loop Closures}
\label{sec:hybridloop}

When ORB-SLAM identifies a loop closure between the latest keyframe $Kf_i$ and a historic keyframe in the map, $Kf_h$, a transformation $T_{i}^{h}$ that aligns $Kf_i$ with $Kf_h$ is computed. This yields a pose pair; the uncorrected pose of the new keyframe $\mathbf{P}_{kf}^i$ and the pose of the keyframe after loop correction $\hat{\mathbf{P}}_{kf}^i$. ORB-SLAM's loop closure mechanism proceeds to apply the loop closure and optimize its sparse map as per the original algorithm \cite{orbslam3}.\par

To update the dense surface to reflect the loop closure $\mathbf{P}_{kf}^i \rightarrow \hat{\mathbf{P}}_{kf}^i$, we non-rigidly deform each surfel using a deformation graph $\mathcal{G}$ in the same way as EF \cite{elasticfusion}. $\mathcal{G}$ consists of a set of nodes $\mathcal{G}^n$ sampled from the surfels in $\mathcal{M}$. Each node consists of a position (i.e., the sampled surfel's position), a timestamp (also sampled surfel's timestamp), an affine transformation $\mathcal{G}^{n}_T$ and a set of temporally nearby neighbor nodes. The temporal connectivity of the nodes prevents interference between multiple passes of the same surface at different points in time. To apply $\mathcal{G}$ to $\mathcal{M}$, a set of temporally and spatially nearby influencing nodes is computed for each $\mathcal{M}^s \in \mathcal{M}$. A weighted sum of the $\mathcal{G}^{n}_T$ in these nodes are applied to the surfel to deform it into place.\par

To optimize the $\mathcal{G}^{n}_T$, a set of surface-to-surface correspondences are generated from a loop closure, mapping surface points from the active region to the inactive region of the model. Cost function terms on the distance between correspondences, as well as terms that maximize the rigidity and smoothness of the deformation, ensure that, when applied to the surface, it brings the correspondences together. A pinning term holds the inactive map in place, ensuring that it is the active map that is deformed into the inactive map. Further constraints on the relative position of historical correspondences from previous loop closures prevent new deformations from breaking the old loop closures. The final cost is optimized using Gauss-Newton gradient descent to retrieve the set of deforming affine transforms $\mathcal{G}^{n}_T$. See \cite{elasticfusion} for more details.\par

In order to reflect ORB-SLAM loop closures in the dense model in our system using the deformation graph, we first generate a set of surface-to-surface correspondences from $\mathbf{P}_{kf}^i \rightarrow \hat{\mathbf{P}}_{kf}^i$. To do so, the active surfels $\mathcal{M}^a \in \mathcal{M}$ in view of $\mathbf{P}_{kf}^i$ are computed, as are the timestamps of the inactive surfels in view of $\hat{\mathbf{P}}_{kf}^i$. These timestamps are used during optimization of $\mathcal{G}^{n}_T$ to pin the inactive surfels of $\mathcal{M}$ in place. Each of the $\mathcal{M}_s^a$ generates a constraint consisting of a source point, computed by composing $\mathcal{M}_s^a$, with $\mathbf{P}_{kf}^i$ and a destination point computed by composing $\mathcal{M}_s^a$ with $\hat{\mathbf{P}}_{kf}^i$. These constraints are then used to optimise the deformation in the same way as \cite{elasticfusion}.\par
\begin{table}[!t]
\centering
\begin{tabular}{@{}cccc|cc@{}}
\toprule
\textbf{Sequence}
& \multicolumn{1}{c}{\begin{tabular}[c]{@{}c@{}}\textbf{O}\\\textbf{$t_{rel}(\%)$}\end{tabular}}
& \multicolumn{1}{c}{\begin{tabular}[c]{@{}c@{}}\textbf{H}\\\textbf{$t_{rel}(\%)$}\end{tabular}}
& \multicolumn{1}{c|}{\begin{tabular}[c]{@{}c@{}}\textbf{H$+$L}\\\textbf{$t_{rel}(\%)$}\end{tabular}}
& \multicolumn{1}{c}{\begin{tabular}[c]{@{}c@{}}\textbf{D$3$\cite{d3vo}}\\\textbf{$t_{rel}(\%)$}\end{tabular}}
& \multicolumn{1}{c}{\begin{tabular}[c]{@{}c@{}}\textbf{ORB\cite{orbslam2}}\\\textbf{$t_{rel}(\%)$}\end{tabular}}\\
\midrule
01 & $31.17$ & $24.49$ & $25.34$ & $1.07$ & $1.39$ \\ 
02 & $2.97$ & $3.11$ & $2.38$ & $0.80$ & $0.76$ \\ 
06 & $1.16$ & $1.42$ & $1.34$ & $0.67$ & $0.51$ \\ 
08 & $1.90$ & $2.14$ & $2.11$ & $1.00$ & $1.05$\\ 
09 & $1.85$ & $2.02$ & $1.62$ & $0.78$ & $0.87$ \\ 
10 & $1.24$ & $1.50$ & $1.41$ & $0.62$ & $0.60$\\ 
\bottomrule
\end{tabular}
\caption{\add{\textbf{Results on test set of KITTI odometry}}. Results from ORB-SLAM stereo and a SOTA VO system D$3$VO are shown for comparison. \del{These results show that our system can provide accurate camera tracking with dense depth prediction while building a fused $3D$ surfel model} \add{Though not SOTA, these results show that our system's camera tracking achieves a level of accuracy necessary for building a fused $3D$ surfel model}. It can be seen that using predicted depth leads to accurate RGBD tracking with ORB-SLAM (\textbf{O}). We show the effect of introducing pre-fusion alignment (\textbf{H}) and applying global loop closures to the dense model (\textbf{H$+$L}).}
 \label{table:kitti_accuracy}
\end{table}
Note that $Kf_i$ is not necessarily the current live camera frame. It could therefore fall outside the current active region of the map. If it is outside the active region of the map, then this will hinder the system's ability to generate surface-to-surface constraints. We found that by setting the active time window appropriately ($\delta t = 200 frames$), the system could find sufficient constraints during loop closures.\par 

Closing loops in this way achieves two goals; adjusting the dense surface geometry to stay consistent with the real world; and keeping the dense map consistent with ORB-SLAM's sparse map and camera pose estimates. It also balances the need for this consistency against the computational intensity of correcting dense geometry. Figure \ref{fig:loop} illustrates each of the steps involved in the hybrid loop closure process. \par 
\section{Evaluation}
\label{sec:evaluation}

We show quantitative and qualitative results of our system tested on the KITTI odometry benchmark dataset \cite{kitti_benchmark}. We have evaluated the accuracy of trajectory estimation and surface reconstruction and provide a breakdown down of the computational performance of the system. All processing was done on a machine with an Intel Core $i7-7700K$ CPU, $16$GB of RAM, and an {NVIDIA} GTX $1080$ {Ti} GPU.
\subsection{KITTI - Tracking}

\begin{figure}[!t]
  \centering
\begin{tabular}{cc}
 	\begin{overpic}[width=0.498\columnwidth]{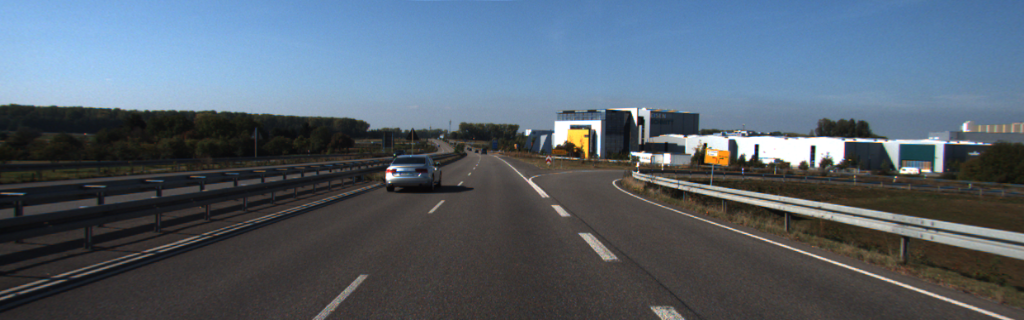}
    \put (0,3) {{$\displaystyle\textcolor{green}{\text{(a)}}$}}
    \end{overpic}
    \begin{overpic}[width=0.498\columnwidth]{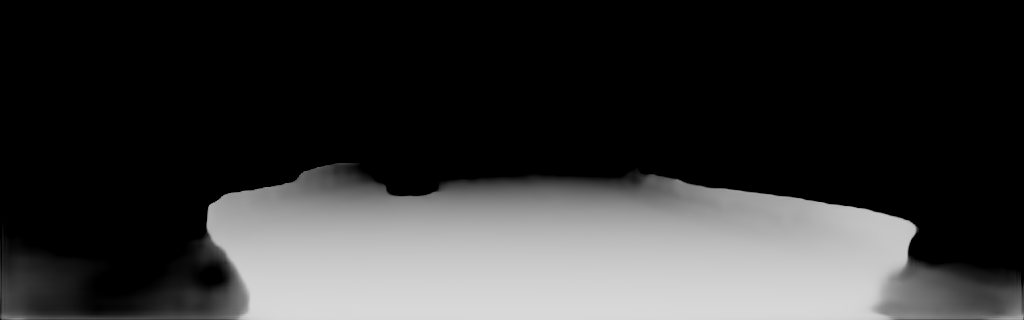}
    \put (0,3) {{$\displaystyle\textcolor{green}{\text{(b)}}$}}
    \end{overpic} \\
    \begin{overpic}[width=0.498\columnwidth]{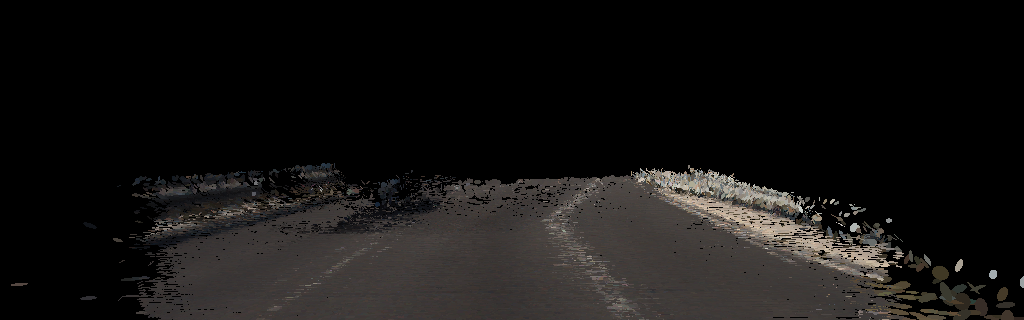}
    \put (0,3) {{$\displaystyle\textcolor{green}{\text{(c)}}$}}
    \end{overpic}
 	\begin{overpic}[width=0.498\columnwidth]{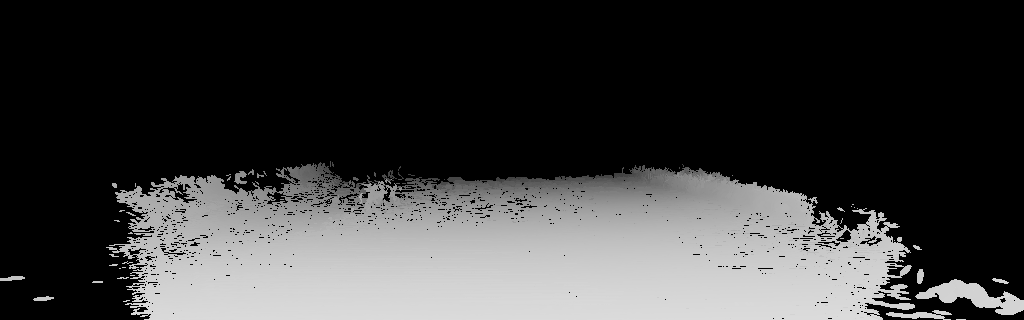}
    \put (0,3) {{$\displaystyle\textcolor{green}{\text{(d)}}$}}
    \end{overpic} \\
    \begin{overpic}[width=0.498\columnwidth]{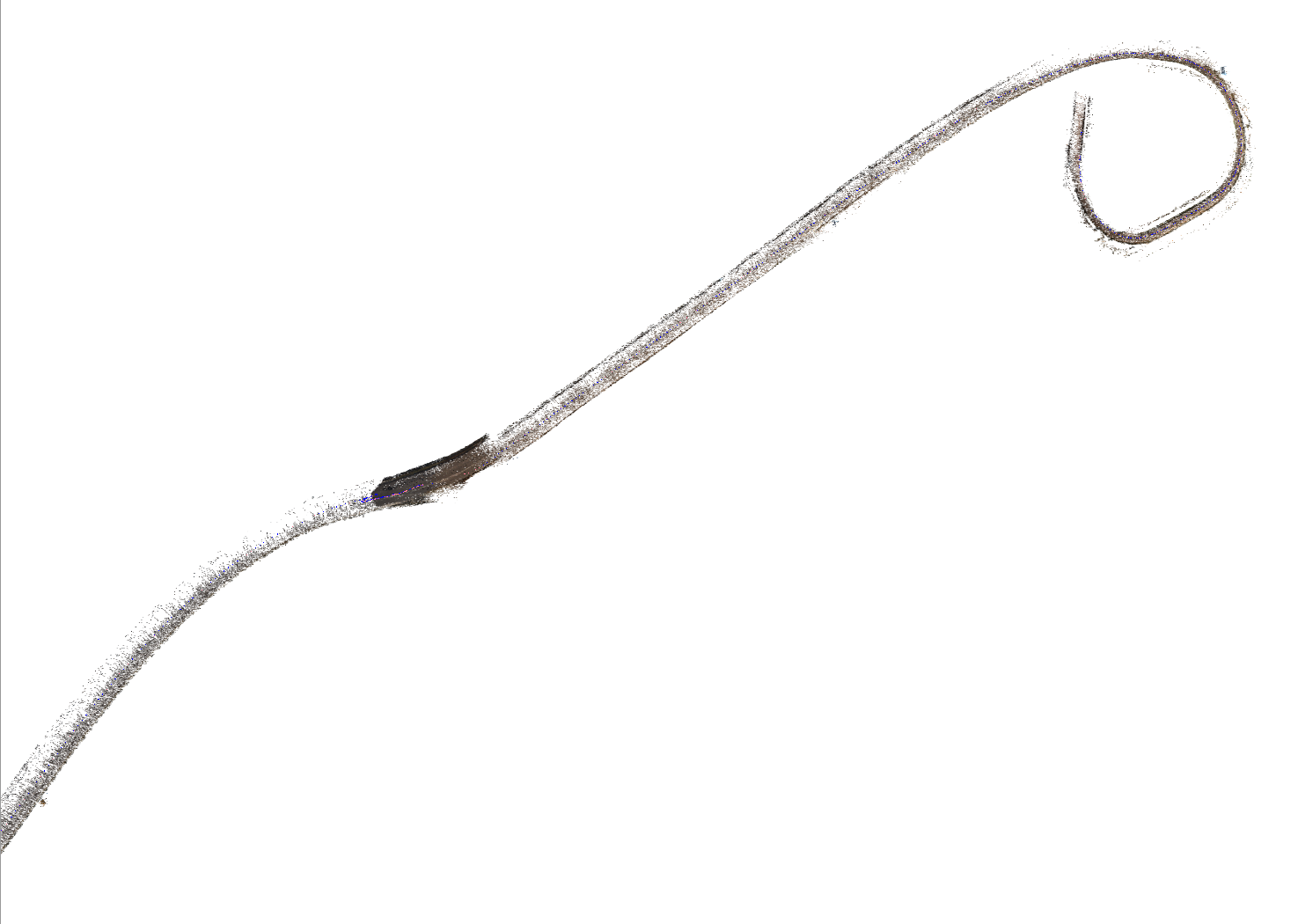}
    \put (0,3) {{$\displaystyle\textcolor{red}{\text{(e)}}$}}
    \end{overpic} 
    \begin{overpic}[width=0.498\columnwidth]{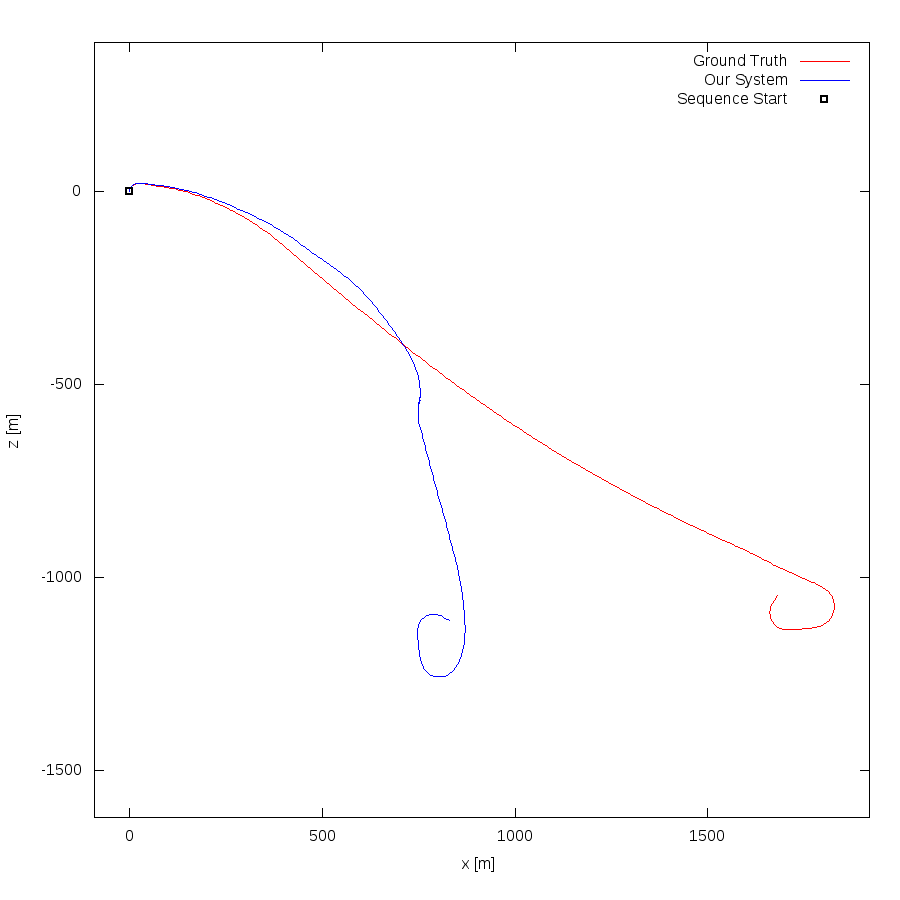}
    \put (0,3) {{$\displaystyle\textcolor{red}{\text{(f)}}$}}
    \end{overpic}
\end{tabular}
\caption{\textbf{Sequence $01$ from the KITTI odometry dataset is challenging for monocular systems}. (a) and (b) show the live color image from the camera and UnRectDepthNet's depth prediction respectively. (c) and (d) show ElasticFusion's corresponding model predictions. (e) shows a portion of the final global model. (f) shows a comparison between the estimated and ground-truth trajctories.}
\label{fig:challenge}
\end{figure}

The KITTI odometry benchmark dataset provides $11$ sequences with ground truth poses. We show results for sequences $01,02,06,08,09$ and $10$. The remainder of the sequences are used during training of the depth prediction network\add{ and hence ommitted from our evaluation}. For each test sequence we use the relative translational error $t_{rel}$ averaged over sequences ranging in length from $100m$ to $800m$ \cite{kitti_benchmark}. Table \ref{table:kitti_accuracy} shows the results alongside a SOTA SLAM system, ORB-SLAM2 \cite{orbslam2}, and D$3$VO \cite{d3vo}, a SOTA VO system. Interestingly, our experiments show that using the dense depth predictions and the RGBD operating mode of ORB-SLAM (\textbf{O}) results in accurate metric-scale camera tracking with a monocular camera. While introducing hybrid tracking (\textbf{H}) increases error w.r.t the ground truth, it allows the current frame to be fused into the model. Introducing hybrid loop closures (\textbf{H$+$L}) helps bring the sparse and dense models back into alignment and reduces global error in the model and trajectory. Sequence $01$ is challenging for our system. Scenes with little structure lead to a degradation in depth prediction and camera tracking. Figure \ref{fig:challenge} shows qualitative results of this sequence.\par
\subsection{KITTI - Surface Reconstruction}
We evaluate surface reconstruction accuracy on several sequences from the KITTI odometry benchmark. As KITTI does not include ground-truth surface models, we compare against models constructed from the Velodyne point clouds that accompany each sequence. To do so, we transform each point cloud in the sequence into a global coordinate system using the corresponding ground truth poses. We then apply a voxel filter to downsample the resultant point cloud. In Table \ref{table:kitti_surface_accuracy}, we show the surface-to-surface mean distance between points in the estimated model and the nearest point in the Velodyne point cloud model. Before computing the score, the two models are rigidly aligned. \del{Due to the sheer volume of data and compute needed for a direct comparison, we compute results for a number of the smaller KITTI sequences.} \del{We note that the Velodyne point clouds contain dynamic objects in the scene, perturbing these results. We intend to recompute the surface accuracy for a final edit of this work using \cite{semantic_kitti} to filter out these dynamic objects.}\add{We use the labels from \cite{semantic_kitti} to filter out dynamic objects from Velodyne point clouds.}\par

\begin{table}[!t]
\centering
\begin{tabular}{@{}cccc@{}}
\toprule
\textbf{Sequence} 
& \begin{tabular}[c]{@{}c@{}}\textbf{O}\\ (m)\end{tabular}
& \multicolumn{1}{c}{\begin{tabular}[c]{@{}c@{}}\textbf{H}\\ (m)\end{tabular}}
& \multicolumn{1}{c}{\begin{tabular}[c]{@{}c@{}}\textbf{H$+$L}\\(m)\end{tabular}} \\
\midrule
02 & 5.95 & 6.40  & \textbf{4.03} \\
06 & 1.23 & 0.97 & \textbf{0.64} \\
08 & 2.70 & \textbf{2.46}  & 2.72 \\
09 & 1.23 & 1.18  &  \textbf{0.71} \\
10 & 0.89 & \textbf{0.79} & 0.82 \\
\bottomrule
\end{tabular}
\caption{\add{\textbf{Surface Accuracy of our system on KITTI.} Sequence $01$ has been ommitted due to poor tracking performance (see Table~\ref{table:kitti_accuracy})}}
\label{table:kitti_surface_accuracy}
\end{table}
\begin{figure}[!t]
   \centering
   \includegraphics[width=\columnwidth]{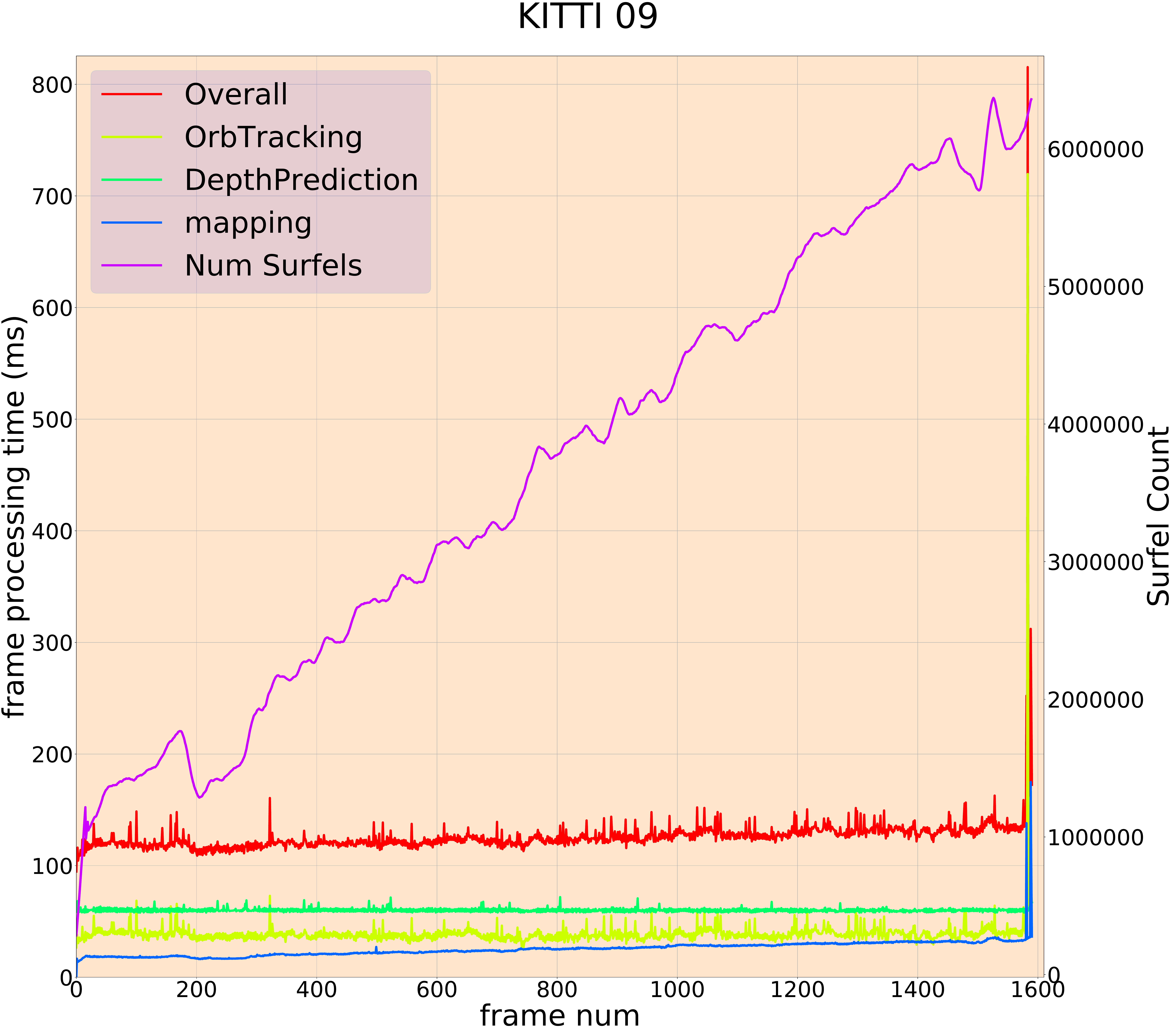}   
   \caption{\add{\textbf{Breakdown of time taken by our system} to process each frame in sequence $09$ the KITTI odometry benchmark \cite{kitti_benchmark}. Our systems falls just outside of true real-time rates of $10hz$ for KITTI. In future work we hope to utilize hardware with half precision support to improve system run-time.}}
   \label{fig:timings}
\end{figure}
\begin{table}[!t]
\centering
\begin{tabular}{@{}cccc@{}}
\toprule
\textbf{Operation} 
& \textbf{Median($ms$)}
& \textbf{Mean($ms$)}
& \textbf{std. dev.($ms$)} \\
\midrule
ORB-SLAM Tracking & 40.04 & 41.93  &  25.13 \\
Dense Mapping & 30.97 &35.61 & 15.82 \\
Depth & 60.32 & 61.14 & 3.84 \\
Overall & 136.48 &  140.29  &  31.76 \\
\bottomrule
\end{tabular}
\caption{\add{\textbf{Spread of system run-time over test sequences from the Eigen split of the KITTI odometry dataset}}}
\label{table:timings_spread}
\end{table}
\subsection{KITTI - Depth Estimation}
We use the depth estimation setting on the \add{KITTI Eigen split~\cite{depth_eigen}} and report results in Table~\ref{tab:kitti-depth}. In our previous work, UnRectDepthNet \cite{unrectdepthnet}, we outperformed all the previous monocular self-supervised methods. Following best practices, we cap depths at 80\,m. We evaluate using the \textit{Improved}~\cite{uhrig2017sparsity} ground truth depth maps. We further improve the performance by incorporating additional losses\footnote{The loss function improvements will be presented in contemporary work to be published at ICRA 2021.} discussed in Section \ref{sec:depth}. Scale estimation improvements provide an additional improvement in quantitative score, enabling the SLAM pipeline to estimate accurate metric reconstructions.
\begin{table}[!t]
\centering
\setlength{\tabcolsep}{0.2em}
\begin{tabular}{@{}lcccc@{}}
\toprule
\textbf{Method}                  
& Abs$_{rel}$ & Sq$_{rel}$ & RMSE & RMSE$_{log}$ \\ 
\midrule
UnRectDepthNet~\cite{unrectdepthnet}  
& 0.081 & 0.414 & 3.412 & 0.117  \\
$+$ additional loss terms \cite{kumar2021omnidet}             
& 0.071 & 0.334 & 3.218 & 0.101  \\
$+$ improved scale estimation         
& \textbf{0.057} & \textbf{0.298} & \textbf{2.95} & \textbf{0.091}  \\     
\bottomrule
\end{tabular}
\caption{\textbf{Evaluation of depth estimation on the Improved KITTI Eigen split \cite{uhrig2017sparsity}.}}
\label{tab:kitti-depth}
\end{table}
\subsection{System Resource Usage}

\del{To show the computational performance of our system,} \add{In Figure~\ref{fig:timings}} we show a breakdown of the frame processing time for sequence $09$ in the KITTI odometry benchmark \cite{kitti_benchmark}. Mapping time (blue) goes up in accordance with the number of surfels (purple) as reported in \cite{elasticfusion}. The system operates between $8-9hz$. The spike at the end of the sequence is due to a global loop closure. \add{Table ~\ref{table:timings_spread} shows the distribution of run-time performance over the test set of the Eigen split of the KITTI odometry dataset.} \par 
\section{Conclusions}

We presented a hybrid SLAM system that combines dense depth prediction with sparse feature tracking and dense surfel fusion techniques. The system permits live-dense metric reconstructions of outdoor scenes using a monocular camera in automotive scenarios. Sparse tracking provides camera pose estimation capable of operating robustly at vehicle speeds. The resulting poses are used within the dense fusion tracking step to initialize a whole image alignment refinement process. Global consistency in the model is maintained through visual place recognition and pose to pose constraints from the sparse system, which again are passed to the dense fusion algorithm where they are integrated with a deformation graph-based map correction step. Our results show competitive performance with SOTA techniques while providing dense fused surface models. \del{We believe our system to be the first dense monocular fusion based visual SLAM system capable of operating in-vehicle scenarios.} \add{We believe our system to be the first dense monocular fusion-based visual SLAM system quantitatively evaluated on automotive scenarios.} Although the focus of this paper is automotive, the system could easily be adapted to other scenarios by retraining the depth prediction network.\par
\bibliographystyle{IEEEtran}
\bibliography{references}
\end{document}